\documentclass[11pt]{article}
\usepackage{coling2020}
\usepackage{times}
\usepackage{graphicx}
\usepackage{latexsym}
\usepackage{amsmath}
\usepackage{booktabs}
\usepackage{verbatim}
\usepackage[utf8]{inputenc}
\usepackage{CJKutf8}
\usepackage{subcaption}
\usepackage[export]{adjustbox}
\usepackage{algorithmic}
\usepackage{xcolor}
\usepackage{danudefs}
\usepackage{xr}
\usepackage{soul}
\usepackage{url}
\usepackage{algorithm}

\colingfinalcopy

\usepackage[disable]{todonotes} 

\makeatletter
\if@todonotes@disabled

\else

\fi
\makeatother

\makeatletter
\newcommand*{\addFileDependency}[1]{
  \typeout{(#1)}
  \@addtofilelist{#1}
  \IfFileExists{#1}{}{\typeout{No file #1.}}
}
\makeatother

\newcommand*{\myexternaldocument}[1]{%
    \externaldocument{#1}%
    \addFileDependency{#1.tex}%
    \addFileDependency{#1.aux}%
}

\myexternaldocument{supplementary}

\begin{document}

\title{Weakly-Supervised Neural Response Selection from \\an Ensemble of Task-Specialised Dialogue Agents}

\author{$\text{Asir Saeed}^\dagger$ \And $\text{Khai Mai}^\dagger$ \ \ \ \ $\text{Pham Minh}^\dagger$ \\ \qquad  \qquad  \qquad \qquad \qquad ${^\dagger}\text{Alt Inc., Japan}$, ${^*}\text{University of Liverpool, UK}$ \\
\qquad \qquad \qquad \qquad \qquad asirsaeed1@gmail.com, khaimt@aimesoft.com, pham.minh@alt.ai, \\ \qquad \qquad \qquad \qquad \qquad nguyen.tuan.duc@alt.ai, danushka@liverpool.ac.uk \And $\text{Nguyen Tuan Duc}^\dagger$ \And $\text{Danushka Bollegala}^*$ 
}

\maketitle

\begin{abstract}
  Dialogue engines that incorporate different types of agents to converse with humans are popular.
   However, conversations are dynamic in the sense that a selected response will change the conversation on-the-fly, influencing the subsequent utterances in the conversation, which makes the response selection a challenging problem. 
   We model the problem of selecting the best response from a set of responses generated by a heterogeneous set of dialogue agents by taking into account the conversational history, and propose a \emph{Neural Response Selection} method.
   The proposed method is trained to predict a coherent set of responses within a single conversation, considering its own predictions via a curriculum training mechanism. 
   Our experimental results show that the proposed method can accurately select the most appropriate responses, thereby significantly improving the user experience in dialogue systems.
\end{abstract}

\section{Introduction}
    Dialogue systems use a variety of mechanisms to carryout  informative and coherent conversations with their users. 
    These conversations can be achieved through rule-based~\cite{varges-etal-2009-combining}, template-based~\cite{wallace2009anatomy}, retrieval-based~\cite{ji2014information} approaches or in a data-driven manner that teaches the agent to learn from raw conversational data~\cite{vinyals2015neural}. 
    Modern dialogue systems often consist of a diverse ensemble of dialogue agents, each of which is individually capable of holding a conversation on its own.     
    Allowing for multiple agents, gives a dialogue system the flexibility to respond to a broad range of user utterances (hereafter, referred to as the \emph{queries}) because each agent can be specialised for a certain task~\cite{serban2017deep} such as task-oriented dialogues or open-ended chit-chat. 
    However, it is important to carefully select the most appropriate response for a given user input.
   For this purpose, we propose a \emph{Neural Response Selection} (NRS) method to select the most appropriate response from a set of responses provided by a set of dialogue agents participating in a dialogue system. 
   We note that ensemble dialogue systems have been proposed in previous work~\cite{song2016two,serban2017deep} and our goal in this paper is \emph{not} to propose a new ensemble learning dialogue system. 
   Instead, we model this as a \emph{response selection} problem, where given a query and a set of responses produced by an arbitrary number of agents, we must select the most appropriate response for the query\footnote{see Supplementary for a conceptual diagram.}.
    
 Ensemble dialogue systems, while reap the benefits of having multiple agents, make response selection problem a challenging task. 
 As a dialogue system evolves, new agents that provide specialised functionalities will be added to the ensemble as well as obsolete agents will be removed.
 The dialogue system must then take into account the new agents' behaviour to ensure that it is picked for the appropriate queries. 
 On the other hand, end-to-end dialogue systems, while benefit from sharing parameters, would have to be retrained in order to account for any newly added agents. 

   We propose an agent-agnostic NRS model for selecting the best response to a given query from a set of responses generated by an ensemble of dialogue agents, where the selection is done \emph{without} any prior knowledge of the constituent agents, significantly differing from ensemble methods or end-to-end dialogue systems. 
    This is made possible because we consider the raw text utterances returned by each agent and the contextual information from the conversation history when selecting responses. 
    In particular, we do not use agents' properties as input features in our model as we want the learnt NRS model to be agent-agnostic and generalise to novel agents. 
    Instead, we resort to contextualised word embeddings~\cite{peters2018deep,devlin2018bert} as the sole input features in our NRS model. 
    Recent works have shown that contextualised embeddings can model syntactically and semantically relevant information that is more abstract than the representations obtained using word co-occurrence-based methods~\cite{gulordava-etal-2018-colorless}. 
   Self-attentive architectures such as Transformers~\cite{vaswani2017attention} have shown to encode hierarchical and syntactic dependencies, and improve dependency parses~\cite{lin2019open,goldberg2019assessing,tenney2019you}.    
    These contextualised features are highly relevant for response selection as we must pick responses that are coherent with the flow of a conversation.    
    However, contextualised word embeddings are not guaranteed to capture discourse-level information and are inadequate to ensure a non-contradictory coherent conversation. 
    Therefore, we use contextualised word representation of a query, the responses of the agents, and the history of the conversation to learn an NRS model.
     As detailed later in Section~\ref{sec:eval}, the proposed NRS method significantly outperforms a baseline that simply uses the semantic similarity between queries and responses, computed using their contextualised representations, as the criteria for selecting responses.
    
    We train our model on a human-written open domain dialogue dataset, covering a broad range of topics. 
    Although we have agents' responses for every query in this datatset, we do not require human annotations for ranking/rating for individual agent responses for a given query.
    Instead, only a human-written single \emph{gold} response per query is used to train the proposed NRS method.
    This is in contrast to supervised regression/classification approaches to response selection/ranking, where human ratings/rankings are required for all candidate responses for a given query.
    However, in our problem setting we \emph{do not} have such labelled scores and choose to train them in a \emph{weakly-supervised} manner. 
    Our labelled data is loosely coupled with our target problem because the dialogue dataset was written for training a neural conversation model and not a response selection model. 
    Although supervised learning would provide us with accurate labels, weak-supervision allows us to reuse training data annotated for a different, but related tasks thereby obviating the need to re-annotate data for the target task.     
    
    To provide weak-supervision to the response selector, we train a critic that compares an agent's response against the human-written gold response for a given query using contextualised embeddings. 
    The critic scores an agent's response highly if it is semantically similar to the corresponding gold response.
    Considering that there can be multiple surface-level correct paraphrases for a gold response, by using contextualised embeddings, the critic will able to detect semantic-level appropriateness of an agent's response to a given user query.   
    Our proposed NRS method takes in contextualised word representations as the input features and outputs a score for a particular (\textsf{context}, \textsf{query}, \textsf{response}) tuple. 
    The \textsf{context} in our case is the \emph{history} of the conversation (i.e utterances from previous time steps) making this a prediction for a future response in time, thereby implicitly capturing a recurrent structure.        
    Although during training time, a sequential prediction model can use the human-annotated targets as the history, this is not available at test time and the model must rely upon its own past predictions, which are less accurate than the human-annotated targets as the history.
    This issue is known as the  \emph{exposure bias}~\cite{Bengio:2015} and we use a curriculum learning mechanism to overcome it, starting with utterance-level training. 
    At test time, the model must select responses considering the entire conversation in a consistent/coherent manner. 
    Hence, we switch to training on entire conversations. 
    After the conversation-level training phase, we conduct scheduled sampling by switching between gold responses and responses selected by our model using teacher-forced annealing~\cite{Goodfellow-et-al-2016}. 
        During our utterance-level training phase our model behaves like any deep neural regression model, outputting a real number from input features. 
        When we move to conversation-level training, the recurrence is further emphasised because the model is predicting the next score based on a potentially \emph{longer} history, where the time-window being set to the entire length of the current conversation. 
        Our experimental results show that the responses selected by the proposed method outperforms a baseline which uses only state-of-the-art contextualised word representations and agrees with the responses selected by human judges.
    

\section{Related Work}
\label{sec:rel}
    \textbf{Learning to Rank}~\cite{Bollegala:GECCO:2011,Liu:2008,pang2017deeprank,song2018deep} is an important task in Information Retrieval (IR) and has some resemblance to our problem setting. 
    Given a query and a set of candidate documents, first, a relevance ranking function is learned to determine the relevance of a document to a query, which is subsequently used to induce a total ordering for the documents. 
    In contrast,  in our setting we must select the \emph{best} (i.e. top-1) ranked response and do not have any use for the other responses, because we can only return one response to the user for a single query. 
    Therefore, the response selection problem in dialogue systems, albeit has some common properties with learning to rank in IR, has significant and important differences that must be carefully considered.     
    A core assumption in learning to rank is that relevant documents must have high semantic similarity to the queries and, hence, the main effort lies in mapping a query and a document into a joint feature space. 
    However, when we are selecting responses for dialogues that may not be the case. In natural conversations, changing the topic, reacting with surprising answers etc. while semantically not related, can still be considered appropriate as long as the conversation is coherent. 
    Moreover, in open-domain conversation the intent might change during the conversation, but for IR within the same session search intent remains constant. 
    Within a single conversation, the best response selected from a set of responses returned by multiple dialogue agents for a given user utterance will change the direction of the conversation and directly influence subsequent utterances in that conversation. Our approach attempts to overcome these challenges by using the conversation history along with the query and the current response and using contextualised vector representations as input features.

    \paragraph{Ensemble Dialogue Systems:}  \newcite{song2016two} proposed an ensemble method for dialogue systems using a combination of a retrieval-based and neural generation based agents. 
    However, they select the final response based on a post re-ranking scorer, which calculates the relatedness between the query and response pair. 
    They used several different types of features such as  the word overlap ratio, the cosine similarity of pre-trained topic model coefficients, and the cosine similarity of word-embedding vectors for calculating the relatedness.
    On the other hand, we use only contextualised word representations as features, and use a neural network to learn how these features appear in (\textsf{context}, \textsf{query}, \textsf{response}) tuples, without any domain-specific feature engineering. 
    \newcite{serban2017deep} used an ensemble of 22 dialogue agents and selected the best response via reinforcement learning.
     Similar to us, they use conversation history. 
     However, their method requires agent-specific attributes as part of its input features, which makes it difficult to incorporate new agents without retraining. 
     Moreover, they use human-annotated ranking labels and learning policy acts in a supervised manner whereas, we do weak-supervision using induced labels from a dialogue corpus. 
        
    \paragraph{Retrieval-Based Agents:}
    \newcite{wu2016sequential} proposed a response selection method for a retrieval-based dialogue agent that considers contextual information from previous utterances. 
    \newcite{yang2018response} designed a response ranking paradigm for dialogue systems by incorporating external knowledge into the matching process of dialogue context and candidate responses. 
    Previous work has shown that contextualised word vector representations for response selection improves the performance of dialogue engines. 
    \newcite{zhou2018multi} used attention mechanism~\cite{vaswani2017attention} for contextualised word vector representation to score matching responses. 
    They use self-attention (word-level dependencies within a sentence) and cross-attention (attention between context and response) to create a richer representation.
    \newcite{nogueira2019passage} fine-tuned a BERT model for query-based passage re-ranking. 
    Although these works present a response selection task which takes conversation context into account or use contextualised word vector representations, the tasks are different to ours. 
    Our work aims to select the correct response from an array of dialogue agent responses to a query, whereas retrieval-based agents are selecting from candidate responses (human-written) in a corpus, where few of which are related to the current query. 
    This distinction is important because our task poses challenges beyond identifying important word and context information and how they relate to the query, and we must process responses produced by a diverse set of dialogue agents specialised to deal with specific situations.
    
\section{Neural Response Selection}
\label{sec:NRS}

    Our NRS method is composed of two main components: a contextualised vector representation from a pre-trained language modelling network (LM), which outputs a $k$-dimensional vector for every token in the sequence, and a response scoring function, $M$. 
    Let us denote an alternating conversation between a user, $u$, and a dialogue engine $e$ by 
    $c_{0}^{(e)}, c_{1}^{(u)}, \ldots, c_{t-1}^{(e)}, c_{t}^{(u)} = q^{(u)}$, where we use $u$ and $e$ in the superscript of each utterance to denote whether it was uttered respectively by the user or the dialogue engine, and $t$ denotes the current time step. 
    The conversation could be initiated by either the user or the dialogue engine and the former sequence shows a conversation initiated by the dialogue engine ($c_{0}^{(e)}$) and the final utterance is produced by the user, denoted by $q^{(u)}$. 
    We refer to $q^{(u)}$ as the \emph{query} and $\cC_{t} = c_{0}^{(e)}, c_{1}^{(u)}, \ldots, c_{t-1}^{(e)}$ as the \emph{history} of the conversation. 
    Our task is to select the best response, $c_{t+1}^{(e)}$, for $q^{(u)}$ from a set of candidate responses, $\{R_{t}^{(j)}\}_{j=1}^{N}$, produced by the $N$ dialogue agents, where the response from the $j$-th dialogue agent at time step $t$ in response to $q^{(u)}$ is denoted by $R_{t}^{(j)}$.
    
    In addition to the agents' response, in our training data, we have a human-written \emph{golden} response, $R_{t}^{(g)}$, for each $q^{(u)}$. 
    We learn a scoring function, $M(R_{t}^{(j)} ; q^{(u)}, \cC_{t}, \Theta)$, that evaluates the appropriateness of $R_{t}^{(j)}$ as a response to $q^{(u)}$, given the history $\cC_{t}$. 
    Specifically, we implement $M$ as a multi-layer feed-forward neural network parametrised by $\Theta$. ReLU is used as the nonlinear activation function throughout the network. 
    Training is done in three phases (1) utterance phase (Section~\ref{sec:utterance-phase}), 
    (2) conversation phase (Section~\ref{sec:conversation-phase})
     and (3) scheduled sampling phase (Section~\ref{sec:SS-phase}).\footnote{See Supplementary Figure~\ref{fig:arch} for a diagrammatic architecture.}
    
    \subsection{Optimisation}
    \label{sec:optimisation}
    For a given query $q^{(u)}$ and its historical context $\cC_{t}$,
    we compute the hinge loss, $L(R_{t}^{(g)}, R_{t}^{(j)}, q^{(u)}, \cC_{t}; \Theta)$, between a response $R_{t}^{(j)}$, returned by the $j$-th dialogue agent and the golden response $R_{t}^{(g)}$, written by a human annotator and time step $t$ as given by \eqref{eq:loss}.
    {\small
    \begin{align}
        \label{eq:loss}
        L(R_{t}^{(g)}, R_{t}^{(j)}, q^{(u)}, \cC_{t}; \Theta) = \max\big( 0, \delta &+ M(R_{t}^{(g)}, q^{(u)}, \cC_{t}; \Theta) - M(R_{t}^{(j)}, q^{(u)}, \cC_{t}; \Theta) \big)
    \end{align}
    }
    Here, $\delta$ is the margin and is set to $1$ in our experiments.
    We use ADAM~\cite{kingma2014adam} to find the $\Theta$ that minimises the hinge loss over all training instances as described later in Section~\ref{sec:exp}.
    
    \subsection{Contextualised Word Embeddings}
    Pre-trained context-independent word embeddings~\cite{mikolov2013distributed,kiela2018dynamic} are inadequate to capture both the syntax and semantics of the words and how its usage varies in different contexts. 
    Contextualised word embeddings on the other hand assign different embeddings to the same word considering its co-occurrence context~\cite{peters2018deep,devlin2018bert}. 
    They can be learned from functions of the internal states of  deep bidirectional language models using LSTMs~\cite{lstms} or Transformers~\cite{vaswani2017attention}. 
    The combinations of the internal states in these networks trained on large text corpus, allow for very rich word representations. 
    Studies have shown integrating these representations into models show better performance across a broad range of NLP tasks~\cite{peters2018deep,devlin2018bert}. 
    The ability to extract these features is particularly important for response selection because we must consider both syntactic, semantic and contextual information. 
    Our proposed method does not assume any underlying properties of the contextualised embeddings and can be used in principle with any contextualised embeddings as we show using AWD-LSTM~\cite{merity2017regularizing}, BERT~\cite{devlin2018bert} and ELMo~\cite{peters2018deep} in experiments.

\subsection{Utterance Phase}
    \label{sec:utterance-phase}    
    The pseudo-code for the utterance-level phase is presented in Algorithm~\ref{alg:trn}, the input for which is the dialogue dataset $\cD$ and the number of epochs $T$, and the output is the trained model, $M$ with optimal parameters $\Theta$. $\cD$ is arranged as a sequence of conversations of the form $\cC_{t}, q^{(u)}, \{R_{t}^{(j)}\}_{j=1}^{N}, R_{t}^{(g)}$ instances. 
    During the utterance phase, we do not use the agent responses as contexts because the human-written responses ($R_{t}^{(g)}$) are available in the dialogue dataset.
    In other words, the history is set to $\cC_{t} = R_{0}^{(g)}, c_{1}^{(u)}, \ldots, R_{t-1}^{(g)}, c_{t}^{(u)}$.
     NRS predicts scores for each candidate agent response and the gold (human-written) response $R_{t}^{(g)}$ at each time step $t$. 
    The parameter update process is done by backpropagating hinge loss between the gold scores and response scores using \eqref{eq:loss}.
    
 \begin{algorithm}[t]
 \small
\caption{NRS in Utterance Phase}
\begin{algorithmic}[1]
\STATE \textbf{Input}: dataset $\cD$, max iterations $T$
\STATE \textbf{Output}: Response scoring model $M$ with parameters $\Theta$
    \FOR{$i \in {0 \ldots T}$}
        \FOR{$(\cC_{t}, q^{(u)},R_{t}^{(g)}) \in \cD$}
            \FOR{$j \in [1 \ldots N]$}
                \STATE $ \mathrm{input} \gets \left(R_{0}^{(g)},c_{1}^{(u)}, \ldots, R_{t-1}^{(g)}\right), q^{(u)},R_{t}^{(j)},R_{t}^{(g)}$
                \STATE $ \mathrm{loss} \gets L(\mathrm{input}; \Theta)$ \COMMENT{$L$ is defined in \eqref{eq:loss}}
                \STATE $\Theta \gets \mathrm{ADAM}(\Theta, \mathrm{loss})$
            \ENDFOR
        \ENDFOR
    \ENDFOR
    \RETURN $\Theta$
    \end{algorithmic}
\label{alg:trn}
\end{algorithm}

    \subsection{Conversation Phase}
    \label{sec:conversation-phase}
    
    During the utterance phase the model mostly observes the best training samples because the provided history consists of human-written gold responses 
    and as a result, NRS is able to select good responses. 
    However, this approach to training assumes that good quality utterances are always available as the history in a conversation. 
    On the other hand, such gold responses are not available during inference/test time and the conversation history will consist of responses selected by NRS.
    This discrepancy between training and inference can lead to accumulated errors with increasing the number of turns in a conversation.
     In order to simulate the test time scenario during training time, we change the training process from a scheme where the human-annotated dialogue history is always present, towards one which uses the agent generated utterance as the dialogue history for the next utterance. 
    Specifically, during training we use the $j$-th agent's past responses, $R_{0}^{(j)}, \ldots, R_{t-1}^{(j)}$ as the history $\cC_{t}$ provided for that agent instead of the human-written gold responses.

    \subsection{Scheduled Sampling Phase}
    \label{sec:SS-phase}
    
    In the conversation phase, NRS uses agents' responses from previous time steps as the dialogue history. 
    Although this is better than using only gold responses as the history during training, it does take into account the responses selected by the NRS model, which form the actual conversation history.
    To make a gradual transition between using gold responses as the history to using responses selected by the NRS model, we propose a curriculum learning approach where we sample from both types of responses according to a linearly scheduled probability, inspired by scheduled sampling~\cite{Bengio:2015}.
    
    The training procedure for conversation (described in Section~\ref{sec:conversation-phase}) and scheduled sampling phase is summarised in Algorithm~\ref{alg:curr}.
    Given a dialogue dataset, $\cD$, we learn an NRS scoring model, $M$, with optimal parameters, $\Theta$, as follows.
    The first few iterations of the training until reaching a phase cut-off threshold, $T_{0}$, correspond to the conversation phase.
    Here, we use the conversation history (Line 7) consisting of an agent's past responses.
    Next, we move to the scheduled sampling phases (Line 8 and below), where we linearly schedule the mixing factor $p$ in Line 9.
    We sample $r$ from the Uniform distribution in the range $[0,1]$ and apply either conversational phase update (Line 12) or scheduled sampling phase update (Line 14) depending on whether $r \leq q$ or otherwise.
    Because $p$ increases from $0$ to $1$, as the training proceeds we will use responses selected by NRS more during the latter parts of the training, simulating the situation during the test time, thereby reducing the exposure bias.
    

\begin{algorithm}[t!]  
\small
\caption{NRS in Conversation and Scheduled Sampling phases}
\begin{algorithmic}[1]
\STATE \textbf{Input}: Dataset $\cD$, max iterations $T$, phase cut-off $T_{0}( < T)$.
\STATE \textbf{Output}: $M$ with parameters $\Theta$
\FOR{$i \in 0 \ldots T$}
    \FOR{$(\cC_{t}, q^{(u)}, \{R_{t}^{j}\}, R_{t}^g) \in \cD$}
        \FOR{$j \in [1 \ldots N]$}
            \IF {$i < T_{0}$}
                \STATE $\mathrm{input} \gets \left(R_{0}^{(j)},c_{1}^{(u)}, \ldots, R_{t-1}^{(j)}\right), q^{(u)},R_{t}^{(j)},R_{t}^{(g)}$
            \ELSE
                 \STATE $p \gets \frac{i - T_{0}}{T - T_{0}}$  
                \STATE $r \sim \textrm{Uni}(0,1)$
                \IF {$r \leq p$}
                    \STATE $\mathrm{input} \gets \left(R_{0}^{(g)},c_{1}^{(u)}, \ldots, R_{t-1}^{(g)}\right), q^{(u)},R_{t}^{(j)},R_{t}^{(g)}$
                \ELSE
                    \STATE $\mathrm{input} \gets \left(c_{0}^{(e)},c_{1}^{(u)}, \ldots, c_{t-1}^{(e)}\right), q^{(u)},R_{t}^{(j)},R_{t}^{(g)}$
                \ENDIF
            \ENDIF
            \STATE $\mathrm{loss} \gets L(\mathrm{input}; \Theta)$  \COMMENT{$L$ is defined in \eqref{eq:loss}}
            \STATE $\Theta \gets \mathrm{ADAM}(\Theta,\mathrm{loss})$
        \ENDFOR
    \ENDFOR
\ENDFOR
\RETURN $\Theta$
\end{algorithmic}

\label{alg:curr}

\end{algorithm}
    
\section{Experimental Setup}
\label{sec:exp}

\subsection{Response Agents}
    \label{sec:sys}
    
    The dialogue system in our setup follows a hybrid architecture, combining hand-crafted rules with machine learning algorithms to address different types of tasks that appear in natural conversations. 
    Human languages are complex and rule-based agents alone are adequate to converse in open-domain topics. 
    On the other-hand machine learning algorithms, while do not require such rules, do not perform well in topics that are not observed in the training data. 
    The hybrid system architecture we use in our setup has been addressed in prior work and shown to outperform single agent dialogue systems~\cite{serban2017deep}. 
    Each of our agents take a user utterance as the input and outputs natural language response. 
    The system we conduct our study on consists of seven independent response agents that use different strategies and do not share any parameters.
    Further details of the agents are described in the Supplementary Section~\ref{sec:agents}.

    \subsection{Dataset}
    \label{sec:dataset}
    To train and evaluate the proposed NRS method, we require
     (a) a dialogue dataset containing conversations between two parties and
     (b) responses from a set of dialogue agents. 
     In Section~\ref{sec:sys}, we already described the dialogue agents and in this section we describe the dialogue dataset that has been developed for the purpose of training a Japanese commercial dialogue system and was made available for us to conduct our experiments.
    The dataset contains multi-turn Japanese conversations and has been created through crowd sourcing. 
    This dataset was created for the purpose of training an open domain commercial neural conversation model~\cite{vinyals2015neural}. 
    A group of ten annotators who were native speakers of Japanese language were commissioned to write conversations averaging 6 turns on a subject picked from a predefined set of 60 topics including \emph{weather}, \emph{movies}, \emph{travelling}, etc. 
    Special care was given to bring out diverse and natural conversations. 
    While an attribute profile was used for the characters in a conversation, we do not use such features in our model. 
    The dialogue dataset contains 6000 conversations covering all 60 topics with at least 100 conversations per topic with a total of 40000 query/response pairs. 
    This dataset is split into mutually exclusive 80\%, 10\%, 10\% parts respectively of train, validation and test sets. 
    Our validation set is used to tune hyperparameters and we report our results on the test set. 
    All topics are equally covered in our dataset splits. 
    
    From this dialogue dataset we create an (1) \emph{utterance-level} and (2) \emph{conversation-level} datasets. 
    The instances for the utterance-level dataset consist of a user query, the previous two utterances, the gold response and the agent responses. 
    We consider responses from all of the seven agents as described in Section~\ref{sec:sys} in our experiments. 
    Our agents do not only respond to user queries but also have the ability to change the subject or ask a question. 
    This differs from dialogue systems which only give agency to the user. 
    It is also important to note that, similar to our proposed NRS model, certain agents also consider dialogue history when producing responses. 
    During the creation of the utterance-level dataset, we consider human annotated previous utterances as context for the agents which require context
    
    For the conversation-level dataset, every instance consists of an average of six turns and contains responses from every agent at every turn. 
    Because a conversation is typically composed of many turns,  we change the dialogue history to alternate between an agent response and a user query from previous turns instead of human-written ones from the dialogue corpus.
    Japanese language does not use space character as the delimiter for separating words.
    Therefore, as a pre-processing step, we conduct tokenisation using MeCab~\cite{mecab2006}, a tokeniser and morphological analyser for Japanese texts. 
    We do not apply lemmatisation or stemming and use surface-level tokens as the input texts. 
    For experiments with BERT~\cite{devlin2018bert} we tokenise using Byte Pair Encoding (BPE)~\cite{sennrich2015neural} implemented in 
    Sentence Piece\footnote{\url{https://github.com/google/sentencepiece}}.
    By using subwords as tokens, we can overcome the out-of-vocabulary (OOV) issues and work with a smaller vocabulary size, which enables us to speed-up the training process and produce smaller models.
    
    \subsection{Training}
    \label{sec:train}
    
    We conduct experiments with contextualised word representations from AWD-LSTM~\cite{merity2017regularizing}, BERT~\cite{devlin2018bert} and ELMo~\cite{peters2018deep}  that have shown to perform well for a wide range of NLP tasks after pre-training on a large corpus. 
    Because our input consists of multiple utterances, we experiment with concatenation and sum pooling for context, query and response utterance representations. In addition, we use max, average and no pooling (only the last hidden state is used) on vectors obtained at every token in the the sequence. 
    The contextualised word representation models are first pre-trained on Japanese Wikipedia articles\footnote{\url{https://github.com/yoheikikuta/bert-japanese}} and then fine-tuned to our dialogue dataset with a language modelling objective in a pre-processing step, prior to using the vectors obtained from them. 

    The hyperparameters for our contextualised word vector models and the process for fine-tuning them to our dialogue corpus follow the same practices as in prior work~\cite{howard2018universal,peters2018deep,devlin2018bert} and are reported in the Supplementary Table~\ref{tbl:hyper}. 
    The vocabulary size is kept to $60,000$ tokens for AWD-LSTM and ELMo and $32,000$ subtokens for BERT (we use a subword tokeniser for BERT). 
    We set the hidden units of NRS to be $128$ and the number of \{\textsf{Dense} \textsf{Layer},\textsf{ReLU}\} blocks in our network to be $3$. 
    The hidden units and the number of layers for NRS were determined using the validation dataset. 
    We train NRS with ADAM~\cite{kingma2014adam} and choose an initial learning rate of $10^{-3}$. 
    We consider only previous two utterances as the context due to GPU memory limitations, because a longer history gives larger contextualised word vectors, particularly for concatenation pooling experiments.

    We train NRS until convergence such that the accuracy on the validation data no longer increases. 
    We keep our hyperparameters fixed throughout all experiments, only changing the contextualised word representation models and the pooling mechanisms. 
    As described in Section~\ref{sec:NRS}, training is done in three phases (1) Utterance phase, (2) Conversation phase and (3) Scheduled Sampling phase. 
    We use our utterance-level dataset for Utterance phase and conversation-level dataset for Conversation and Scheduled-Sampling phase. 
    The training phases are done progressively and the weights are not re-initialised between phases. 
    Utterance phase training is carried out using Algorithm~\ref{alg:trn} with $T=10$ epochs, and Conversation and Scheduled Sampling phases are carried out using Algorithm~\ref{alg:curr} with $T=20$ epochs and phase cutoff $T_0=10$. 
    We consider the phases as fine-tuning processes in our training curriculum, where we condition our model to deal with test-time environments. 
    
    Taking inspiration from Howard and Ruder~\cite{howard2018universal}, in order to retain previous knowledge and avoid forgetting between training phases, we conduct gradual unfreezing of layers. All layers are unfrozen during Utterance phase. During Conversation and Scheduled Sampling phase, layers $1$ and $2$ are frozen, only performing weight updates on layer $3$ from epochs $1$ through $3$ and all layers from epochs $4$ through $20$. Learning rate is initialised to $10^{-3}$ from epochs $1$ through $3$ and lowered to $10^{-4}$ at the beginning of epoch $4$. We concatenate the last hidden states of context, query and response utterances and use it as the input to our model. 
    Although we experimented with concatenation, sum and max pooling, we report results only for the concatenation pooling as it produced the best results.

    \section{Evaluation}
    \label{sec:eval} 
    
    We evaluate our model in an ``oracle'' setting, where we know the correct response for past turns in a given conversation. 
    The correct response labels for our validation and test set are labelled by two human evaluators. The Cohen's kappa score~\footnote{\url{https://scikit-learn.org/stable/modules/generated/sklearn.metrics.cohen_kappa_score.html}} between two annotators is $0.86$ showing strong agreement.
    An evaluator picks the best response given a set of context, query, agent response triples. 
    We do not provide the golden response or information on the agents to the evaluator. 
    We report accuracy scores of our test dataset on this evaluation. 
    A response selected by a response selection method is evaluated to be correct only if it has also been selected by the evaluator.
    If multiple responses are marked by the evaluator as accurate responses, then if at least one of those responses are selected by a response selection method, then we would consider it to be a correct selection. 
    It is possible that all agents returned equally inaccurate responses hence, the human evaluator was unable to pick one. 
    Where no agent responses were selected by the evaluator, we include the golden response as a candidate for selection by the response selection method. 
    We reason that it would not be a fair evaluation to ask the model to select a response among equally inaccurate responses as we do not want our evaluation criteria to be a function of the performance of individual agents. 
    For these instances, selecting the golden response is considered to be an accurate selection.
    It is important to note that we are not evaluating the quality of the responses from agents, rather the accuracy of the selection, the upper bound of which will be 100\% in the case of a model that can make perfect selection.    
    We require a dataset where responses from multiple agents for a single query must be available. Unfortunately, none of the publicly available benchmark datasets for dialogues satisfy this requirement. 
    Therefore, we cannot use existing benchmarks for evaluating a neural response selection method and only compare against our dataset. 

\begin{table*}
\centering
\scalebox{0.7}{
\begin{tabular}{ l  l l l l l l l l l }
\toprule
\multicolumn{1}{c}{Method} & \multicolumn{9}{c }{Contextualised Word Vector Type} \\ 
 & \multicolumn{3}{c}{\textbf{AWD-LSTM}} & \multicolumn{3}{c}{\textbf{BERT}} & \multicolumn{3}{c}{\textbf{ELMo}} \\ 
 \cmidrule(r){2-4} \cmidrule(r){5-7} \cmidrule(r){8-10} \\
 & U.Phase & C.Phase & SS.Phase & U.Phase & C.Phase & SS.Phase & U.Phase & C.Phase & SS.Phase \\ \midrule
Cosine-Similarity (Query,Response) & $39.72$ & - & - & $39.72$ & - & - & $39.72$ & - & - \\ 
NRS (Query,Response) & $47.78$ & - & - & $54.80$ & - & - & $40.21$ & - & - \\ 
Cosine-Similarity (Context,Query,Response) & $38.88$ & $27.77$ & $27.77$ & $38.88$ & $27.77$ & $27.77$ & $38.88$ & $27.77$ & $27.77$  \\
NRS (Context,Query,Response) & $47.49$ & $49.16$ & $52.40$ & $64.00^\ast$ & \textbf{$60.40^\ast$} & $61.94^\ast$ & \textbf{$65.40^\ast$} & $55.00$ & \textbf{$62.80^\ast$} \\ \bottomrule
\end{tabular}}
\caption{Human evaluation of models across phases on the test dataset, $^\ast$ denotes statistical significance}
\label{tbl:human_eval}
\end{table*}

    \section{Results}
    \subsection{Quantitative Analysis}
    We report performance of NRS in Table~\ref{tbl:human_eval} across Utterance (U.Phase), Conversation (C.Phase) and Scheduled Sampling (SS.Phase) phases and architectures. 
   As a baseline, we use Cosine-Similarity~\cite{song2016two,zhou2018multi,nogueira2019passage}, which selects the agent response $R_{t}^{(j)}$ that is closest to the query, measured using the contextualised sentence embeddings for the agent responses and the query.
   Cosine-Similarity between two vectors $\vec{x_1}, \vec{x_2}$ is defined as, $\dfrac{\vec{x}_1\T \vec{x}_2}{\max(\norm{\vec{x}_1}\norm{\vec{x}_2}, \epsilon)}$,where $\epsilon$ is $1e^{-8}$ to avoid division by zero\footnote{\url{https://pytorch.org/docs/stable/nn.html#torch.nn.CosineSimilarity}}. Specifically, averaged pooling over individual contextualised word embeddings in a sentence is used as the sentence embedding for that sentence.
   We denote this baseline as \textbf{Cosine-Similarity(Query, Response)}.
   To consider context within this baseline, denoted as \textbf{Cosine-Similarity(Query, Response, Context)}, we compute the sentence embeddings for each utterance in the history and average it with the query embedding.
   Next, we find the agent response that is closest to this vector according to the cosine similarity. 
   Any improvement over cosine similarity baseline shows the effect of learning a response selection model on top of input representation via contextualised embeddings. 
   Therefore, cosine similarity is an informative and useful baseline to comparing against. 
   Moreover, cosine similarity has been used extensively in NLP for text matching~\cite{zhelezniak-etal-2019-correlation}, it is unsupervised and fast to compute.
    
     NRS achieves $15.08\%$ improvement over the \textbf{Cosine-Similarity(Query, Response)} baseline
     and a $26.52\%$ improvement over the \textbf{Cosine-Similarity(Context, Query, Response)} baseline in the Utterance phase. 
     We see increase in performance for all vector representations with our proposed method. 
     Using ELMo gives the highest accuracy of $65.40\%$ in the Utterance phase and $62.80\%$ in the Scheduled Sampling phase, whereas BERT has the highest accuracy of $60.40\%$ in the Conversation phase. 
     According to the Clopper-Pearson confidence intervals~\cite{Clopper-Pearson} at $p < 0.05$ level, our best model, ELMo, 
     statistically significantly outperforms the Cosine-Similarity baseline. 
     The baseline model performance is $39.72\%$ regardless of contextualised vector representation, and drops respectively to $38.88\%$ 
     and $27.77\%$, when using the context in the U.phase and  C/SS.phases.
     Overall, the accuracy decreases across contextualised word representations in the conversation phase. 
     In this phase, the history consists of the less accurate agents as opposed to gold responses, decreasing the model accuracies. 
     We recover from this in our scheduled sampling phase where we sample from both the gold responses and the model selected responses. 

    \subsection{Qualitative Analysis}
    We conducted a qualitative analysis on the responses produced by the different agents (Agent 1-3), the best response selected by the human evaluator and that by the proposed NRS method. 
  Due to the limited availability of space, we show the responses in Supplementary Table~\ref{tbl:output}.
    
    
     In the first example, we see that although an agent returns the golden response, the human evaluator prefers a different one. 
     In this example, NRS managed to select the the evaluator's selection (an agent response) rather than the golden response from the dialogue corpus. 
     This shows that the model is using the pre-trained contextualised word embeddings and the feedback from hinge loss, to gain an understanding of a dialogue in general as opposed to over-fitting to the training data and always preferring the golden response. 
     The second example on the other hand is much easier as all responses are equally accurate. 
     Since agents maybe capable of prompting the user, changing topics etc. the model must be adaptable to these situations and select appropriately. 
     A model trained on always picking the golden response will not perform reliably at test-time as the agent responses can diversify while being equally (or more) appropriate than the gold. 
     A strict feedback policy like cross entropy or mean-squared error (without annotated scores) would encourage the model to simply be a binary classifier between the golden and agent responses, making it note very useful at test-time  where the golden responses maybe unavailable. 
     Upon inspecting the output we see that the hinge loss avoids this problem  because the criteria for a good score is enforced by a weakly-supervised signal.

\section{Conclusion}
    We proposed a neural response selection method to select the best response from an ensemble of dialogue agents. 
    We used contextualised word embeddings to represent conversation history, query and response utterances as input features to our model. 
    We showed that response selection can be learned independently of the dialogue system using a feed-forward neural network. 
    We use a curriculum training mechanism, moving from Utterance, Conversation to Scheduled Sampling phases in order to make it robust to real-world settings. 
    We empirically show that our model is able to select relevant responses and outperforms cosine-similarity inspired baselines. 
    

\section*{Supplementary Material}

\subsection{Descriptions of the Dialogue Agents}
\label{sec:agents}

\begin{figure}[h!]
        \centering
        \includegraphics[width=0.7\textwidth]{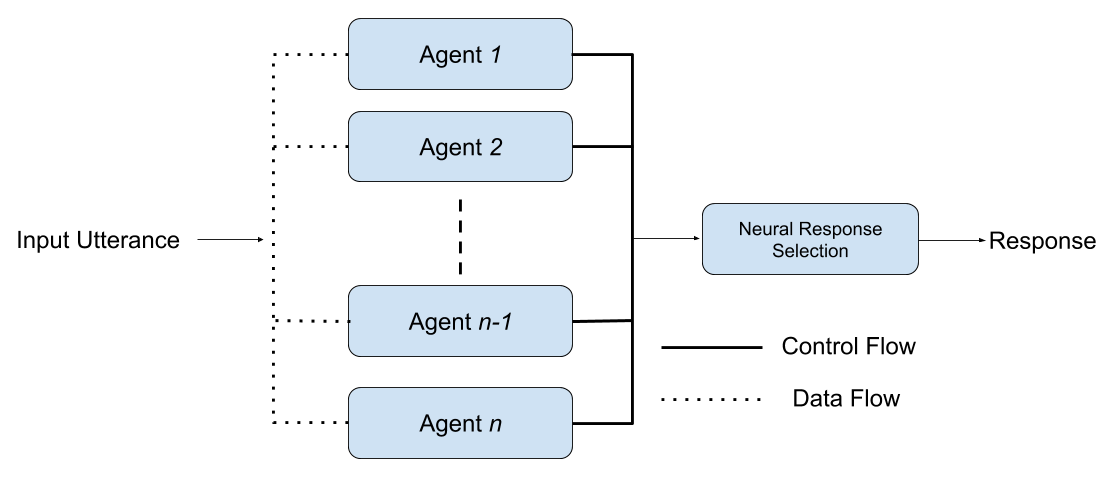}
        \caption{Outline of the proposed neural response selection method}
        \label{fig:outline}
    \end{figure}

In this section we provide the details of our hyperparameters in Table~\ref{tbl:hyper}, outline figures of our setup in Figure~\ref{fig:outline} and architecture in Figure~\ref{fig:arch} and output samples from our dataset in Table~\ref{tbl:output}. We also provide details of the agents used in our experimental setup.

    \begin{description} 
    \item{\textbf{Template-based Agent 1:}}
    The template-based dialogue agent use a set of AIML (artificial intelligence markup language) templates to produce a response given the dialogue history and user utterance. The template based agent consists of rules written in AIML to support frequently asked queries. For example, greeting-type queries, queries on the attributes of the agent (name, age etc) can easily be supported by using string matching for patterns.    
    \item{\textbf{Retrieval-based Agent 2:}}
    The retrieval-based agent retrieves related responses from queries in the corpus that are similar to the given query. We would not generate a new response, but select the most suitable response (originally made to other queries) as reply to the current query. This is done using Word2Vec embeddings to find matching queries in our training dialogues with the highest cosine similarity.    
    \item{\textbf{Neural Generation Agents 3 and 4:}}
    The neural generation agent is a sequence-to-sequence~\cite{vinyals2015neural} model trained on our dialogue corpus with query/response pairs. We use two architectures of sequence-to-sequence models: a hierarchical model~\cite{serban2017hierarchical} that uses context information from the previous utterances in the conversation and a model that uses the copying mechanism~\cite{gu2016incorporating} to create two versions of neural generation dialogue agents.    
    \item{\textbf{Question Generation Agents 5, 6 and 7:}}
    Apart from the agents that produce responses to user queries, we also have agents that prompt the user with questions. We do this because natural conversation is not a question-answer turn based scenario. Questions can have follow up questions and answers can be followed by statements. To generate questions we also use two other neural generation agents that have the same architecture as the ones mentioned before except during training we swap the query/response pairs to generate questions as responses for input utterances. We also use retrieval-based question generation agent where instead of looking for similar replies, we look for similar queries in our dialogue corpus, given a user utterance.
    \end{description}
    
    \newpage
 \subsection{System overview and Network Architecture}
 \label{sec:system}

\begin{figure}[h!]
    \centering
    \includegraphics[width=0.7\textwidth]{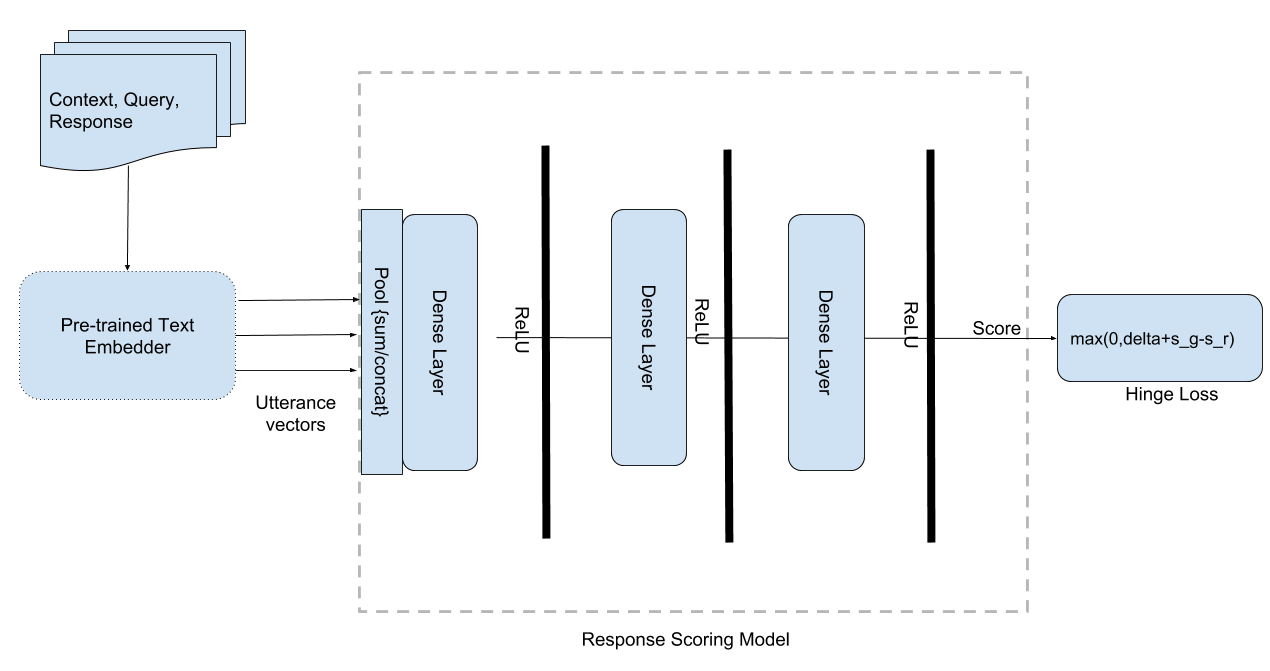}
    \caption{Network Architecture of the Proposed Neural Response Selection Method}
    \label{fig:arch}
\end{figure}

\subsection{Hyperparameter Values}
\label{sec:hyper}

\begin{table}[h!]
\small
\centering
\scalebox{0.9}{
\begin{tabular}{p{2cm} p{2cm} p{1.2cm} p{1cm}}
\toprule
Architecture & Word embedding Size & Hidden layer size & Num. Layers \\
 \midrule
\textbf{AWD-LSTM} & 400 & 1150 & 3  \\
\textbf{BERT} & 300 & 768 & 12  \\
\textbf{ELMo} & 400 & 1024 & 3  \\
\textbf{NRS} & - & 128 & 3  \\
\bottomrule
\end{tabular}}
\caption{Hyperparameters in NRS.}
\label{tbl:hyper}
\end{table}

\newpage
\subsection{Qualitative Analysis Results}
\label{sec:qualitative}

\begin{table*}[h!]
\centering
\scalebox{0.9}{
\begin{tabular}{p{1cm}|p{1cm}|p{2cm}|p{2cm}|p{2cm}|p{2cm}|p{2cm}|p{3cm}}
  \toprule
 \multicolumn{1}{p{1cm}|}{\textbf{NRS}} &
 \multicolumn{1}{p{2.5cm}|}{\textbf{Human Evaluator}} &
 \multicolumn{1}{p{2cm}|}{\textbf{Context}} & 
 \multicolumn{1}{p{2cm}|}{\textbf{Query}} & 
 \multicolumn{1}{p{2cm}|}{\textbf{Agent 1}} &
 \multicolumn{1}{p{2cm}|}{\textbf{Agent 2}} &
 \multicolumn{1}{p{2cm}|}{\textbf{Agent 3}} &
 \multicolumn{1}{p{3cm}}{\textbf{Gold Response}}\\
 \midrule
 Agent 1 & Agent 1 & \begin{CJK}{UTF8}{min}[私はSFが好きで、ガンダムが特に好きかな; 私も、新しい方のガンダムを見ていました]\end{CJK} \newline [I like science fiction, especially Gundam; I was watching the new Gundam show] & \begin{CJK}{UTF8}{min}長いシリーズだから、何から見たかで年齢がばれてしまいますね\end{CJK} \newline It's a long running show, sharing which ones I've seen could spill by age & 
 \begin{CJK}{UTF8}{min}何がきっかけで始めたんですか？\end{CJK} \newline Was there a reason behind starting? & 
 \begin{CJK}{UTF8}{min}そういえば、最近は、宇宙戦艦ヤマトのような昔のアニメがリメイクされていますね\end{CJK} \newline Oh by the way, recently they've started remaking old anime like Battleship Yamato. &
 \begin{CJK}{UTF8}{min}確かに！\end{CJK} \newline Oh that's right! &
 \begin{CJK}{UTF8}{min}そういえば、最近は、宇宙戦艦ヤマトのような昔のアニメがリメイクされていますね\end{CJK} \newline Oh by the way, recently they've started remaking old anime like Battleship Yamato.\\
 \hline
 Agent 2 & Agent 1,2,3 & \begin{CJK}{UTF8}{min}[バイキングはお好きですか; はい、たまに行きます。]\end{CJK} \newline [Do you like all-you-can-eat places? Yes, I go sometimes.] & 
 \begin{CJK}{UTF8}{min}元を取る自信はありますか？。\end{CJK} \newline Do you think its worth the price? &
 \begin{CJK}{UTF8}{min}あります。\end{CJK} \newline Yes I do. &
 \begin{CJK}{UTF8}{min}そうですね。\end{CJK} \newline Yes, that's correct &
 \begin{CJK}{UTF8}{min}特にありません。\end{CJK} \newline Not really.  &
 \begin{CJK}{UTF8}{min}あれもこれもと手は伸びるんですけど、元が取れているとは思いません。\end{CJK} \newline I'm trying my best to make it worthwhile, but doesn't feel like I managed to\\
\bottomrule
\end{tabular}}
\caption{Output Samples from the Response Selection Dataset}
\label{tbl:output}
\end{table*}

\bibliography{coling2020}

\end{document}